\documentclass{winnower}
\usepackage{graphicx}
\usepackage{float}
\usepackage{graphicx,caption}
\usepackage{hyperref}
\hypersetup{colorlinks,urlcolor=blue}
\usepackage[symbol]{footmisc}
\usepackage[all]{nowidow}
\begin{document}

\makeatletter
\renewcommand\@biblabel[1]{#1.}
\makeatother

\newcommand\blfootnote[1]{%
  \begingroup
  \renewcommand\thefootnote{}\footnote{#1}%
  \addtocounter{footnote}{-1}%
  \endgroup
}

\title{\textbf{Humans can decipher adversarial images}
}

\author[ ]{Zhenglong Zhou \& Chaz Firestone\footnote{To whom correspondence should be addressed:  \href{mailto:chaz@jhu.edu}{chaz@jhu.edu}.}}
\affil[ ]{Department of Psychological \& Brain Sciences, Johns Hopkins University}

\date{}

\maketitle

\blfootnote{\\ Version: 12/27/18 (manuscript submitted for publication)}

\vspace*{-1cm}

\begin{abstract}

\noindent {\fontseries{b}\selectfont Does the human mind resemble the machine-learning systems that mirror its performance? Convolutional neural networks (CNNs) have achieved human-level benchmarks in classifying novel images; these advances support technologies such as autonomous vehicles and machine diagnosis, and are even candidate models for human vision itself. However, unlike humans, CNNs are ``fooled" by adversarial examples---nonsense patterns that machines recognize as familiar objects, or seemingly irrelevant image perturbations that nevertheless shift the machine's classification. Such bizarre behaviors challenge the promise of these new advances---but do human and machine judgments fundamentally diverge? Here, we show that human and machine classification of adversarial images are robustly related: In 8 experiments on 5 prominent and diverse adversarial imagesets, human subjects correctly anticipated the machine's preferred label over relevant foils---even for images described as ``totally unrecognizable to human eyes". Human intuition is thus a surprisingly reliable guide to machine (mis)classification---with consequences for minds and machines alike.
}\\

\end{abstract}

\section*{Introduction}

\vspace*{0.2cm}

\noindent How similar is the human mind to the machines that can behave like it? After decades of failing to match the recognitional capabilities of even a young child, machine vision systems can now classify natural images with accuracy rates that match adult humans (Krizhevsky, Sutskever, \& Hinton, 2012; Russakovsky et al., 2015). The success of such models, which have their basis in biologically inspired Convolutional Neural Networks (CNNs; LeCun, Bengio, \& Hinton, 2015), has been exciting not only for the practical purpose of developing new technologies that could automate image recognition (e.g., screening baggage at airports, reading street signs in autonomous vehicles, or diagnosing radiological scans), but also for better understanding the human mind itself. Recent work, for example, has found that CNNs can be used to predict patterns of neural firing, regional activation, and behavior in humans and non-human primates, leading to speculation that the mechanisms and computational principles underlying CNNs may resemble those of our own brains (Cichy et al., 2016; Greene \& Hansen, 2018; Jozwik et al., 2017; Kriegeskorte, 2015; Kubilius et al., 2016; O'Connell \& Chun, 2018; Peterson et al., 2016; Yamins \& DiCarlo, 2016).\\

\begin{figure*}
\begin{center}
\includegraphics[scale=0.21]{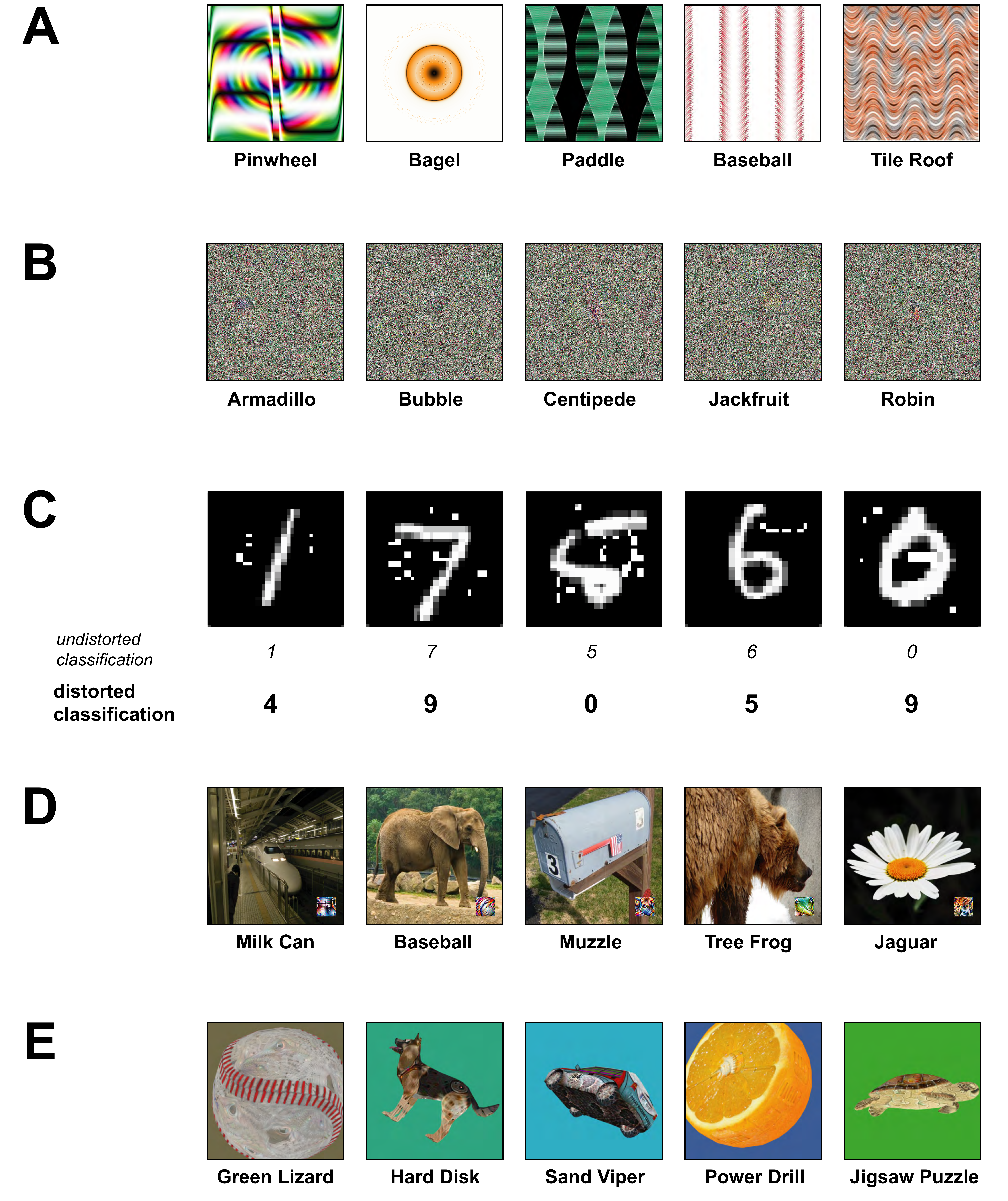}
\captionsetup{width=.85\linewidth,font=footnotesize}
\caption
{Examples of adversarial images that ``fool" Convolutional Neural Nets trained to classify familiar objects, with labels indicating the machine's classifications. (A) Indirectly encoded ``fooling" images (from Nguyen et al., 2015). (B) Directly encoded ``fooling" images (from Nguyen et al., 2015). (C) Perturbed adversarial images that cause the machine to classify one kind of digit as another (from Papernot et al., 2016). (D) The LaVAN attack (Karmon et al., 2018) can cause a machine to misclassify a natural image even when the noise is localized to a corner of the image. (E) ``Robust" adversarial images (Athalye et al., 2017) are 3D objects that are misclassified from multiple viewpoints (and can even be physically produced in the real world).}
\label{fig:Figure 1}
\end{center}
\end{figure*}

\noindent However, such models of object classification differ from humans in a crucial, alarming, and even bizarre way: They are vulnerable to attack by an ``adversary" --- a second model trained to produce images that ``fool" the image-recognition model into misclassifying (Athalye et al., 2017; Karmon, Zoran, \& Goldberg, 2018; Nguyen, Yosinski, \& Clune, 2015; Papernot et al., 2016; Szegedy et al., 2014; for a review, see Biggio \& Roli, 2018). Two especially striking classes of such adversarial images are \textit{fooling} images and \textit{perturbed} images (Fig.\ \ref{fig:Figure 1}). Fooling images are otherwise meaningless patterns that are classified as familiar objects by the machine. For example, a collection of oriented lines might be classified as a ``baseball", or a colorful television-static-like image might be called an ``armadillo". Perturbed images are images that would normally be classified accurately and straightforwardly (e.g., an ordinary photograph of a daisy, or a handwritten number 6) but that are perturbed only slightly to produce a completely different classification by the machine (e.g., a jaguar, or a handwritten number 5).  \\

\vspace*{0.05cm}

\noindent Adversarial images mark an ``astonishing difference in the information processing of humans and machines" (Brendel et al., 2018), and fundamentally challenge the promise of these new approaches. First, and more practically, adversarial images could enable malicious attacks against machine vision systems in applied settings (e.g., changing how an autonomous vehicle reads a street sign; Eykholt et al., 2018). Second, and more theoretically, the fact that such bizarre images are straightforwardly classified as familiar objects by the machine seems to reveal how alien the CNN's underlying processing must be (Griffiths, Abbott, \& Hsu, 2016; Guo et al., 2016; Rajalingham et al., 2018; Yamins \& DiCarlo, 2016), which in turn diminishes their utility as avenues for understanding the human mind.  \\

\vspace*{0.05cm}

\noindent A primary factor that makes adversarial images so intriguing is the intuitive assumption that a human would not classify the image as the machine does. (Indeed, this is part of what makes an image ``adversarial" in the first place, though that definition is not yet fully settled.) However, surprisingly little work has actively explored this assumption by testing human performance on such images, even though it is often asserted that adversarial images are ``totally unrecognizable to human eyes" (Nguyen et al., 2015, p.427). At the same time, it has never been clear under which conditions human and machine performance might be usefully compared, and you may informally observe that at least some adversarial images ``make sense" once you are told the label assigned by the machine (as in Fig.\ \ref{fig:Figure 1}). This raises an intriguing question: Could humans decipher such images by predicting the machine's preferred labels? If so, this might suggest a greater overlap between human and machine classification than adversarial images seem to imply, and could even point to human intuition as a piece of the more practical puzzle of defending against such attacks. \\

\vspace*{0.05cm}

\noindent To address this question, we introduce a ``machine-theory-of-mind" task that asks whether humans can infer the classification that a machine-vision system would assign to a given image.  We acquired images produced by several prominent adversarial attacks, and displayed them to human subjects who were told that a machine had classified them as familiar objects. The human's task was to ``think like a machine" and determine which label was generated for each image. (For a related task with natural images, see Chandrasekaran et al., 2017.) We conducted eight experiments using this task, probing human understanding of five different adversarial imagesets (from Athalye et al., 2017; Karmon et al., 2018; Nguyen et al., 2015; Papernot et al., 2016). Importantly, none of these images was created with human vision in mind (cf. Elsayed et al., 2018) --- they were simply generated to fool a machine-vision system into misclassifying an image. \\

\vspace*{0.05cm}

\noindent Across these 8 experiments --- covering a diverse array of adversarial attacks, as well as several variations on the core experimental design --- we find that human subjects can anticipate the machine's classifications of adversarial stimuli. We conclude that human intuition is a far more reliable guide to machine (mis)classification than has typically been imagined, and we discuss this implications of this result for comparisons between human and machines in the context of cognitive science and artificial intelligence. 

\vspace*{0.25cm}

\section*{Results}

\subsection*{Experiment 1: ``Fooling'' images with foil labels}

\vspace*{0.2cm}

Our first experiment administered the machine-theory-of-mind task using 48 ``fooling" images that were produced by an evolutionary algorithm to confound a highly influential image-recognizing CNN, AlexNet (Krizhevsky et al., 2012), which classified them as familiar objects such as \textit{pinwheel} and \textit{bagel}. (There is evidence that adversarial images for one CNN often transfer to others; Tram\`er et al., 2017.)\\

\begin{figure*}[!ht]
\centering
\includegraphics[scale=0.17]{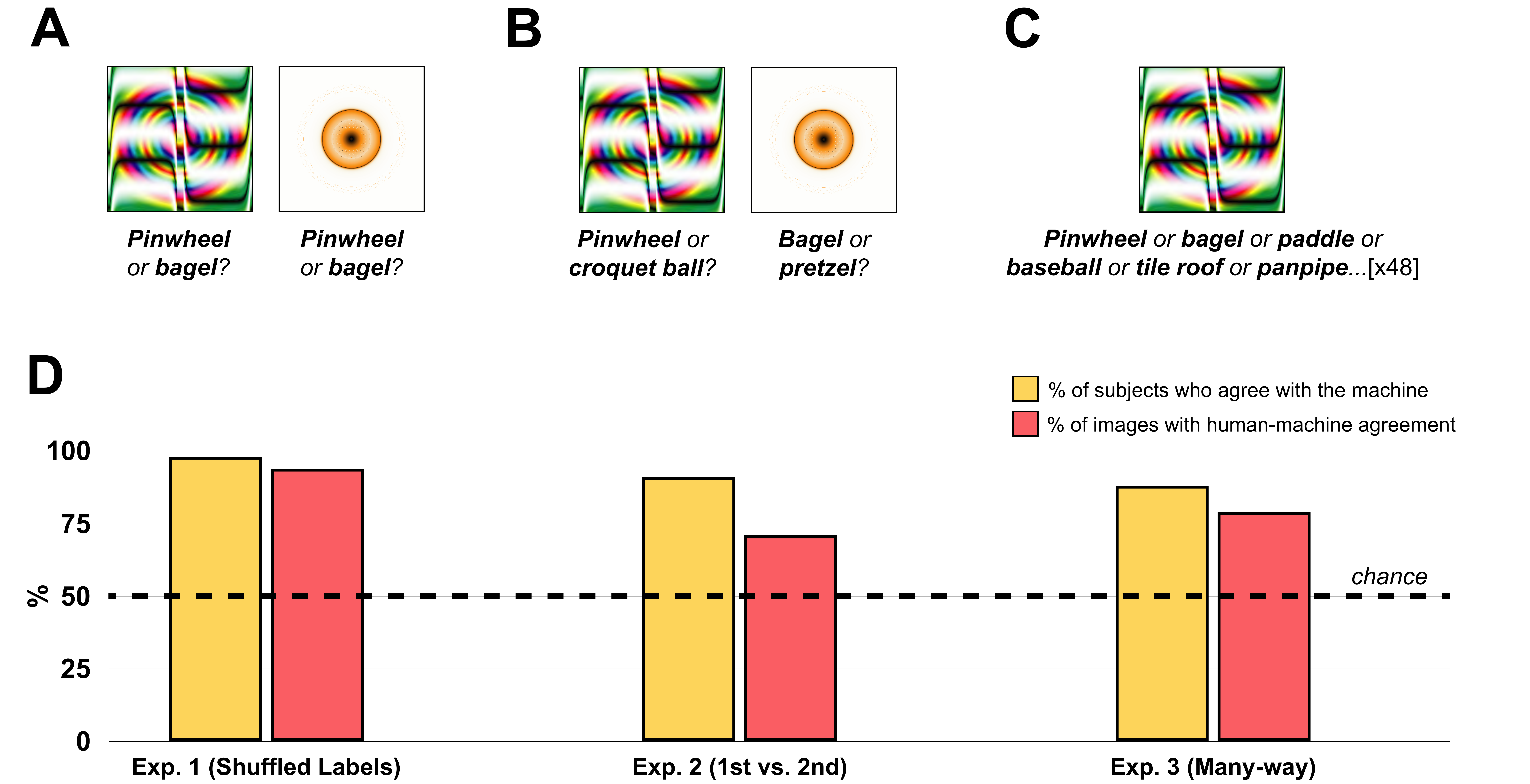}
\captionsetup{width=.85\linewidth,font=footnotesize}
\caption
{Forced-choice classification with indirectly encoded fooling images. (A) In Experiment 1, 200 subjects saw one fooling image at a time (48 images total), and chose between two candidate labels: The machine's choice for that image, and a random label drawn from the other images in the imageset. (B) In Experiment 2, 200 subjects chose between the machine's first-choice classification and its second-choice classification. (C) In Experiment 3a, 200 subjects saw the same images as before, but with all 48 labels visible at once. (D) In all 3 experiments, most subjects agreed with the machine more often than would be predicted by chance responding (yellow bars), and most images showed human-machine agreement more often than would be predicted by chance responding (red bars). Given that raw classification accuracy in human subjects will be modulated by factors such as attention, motivation, engagement with the task, time spent studying an image, etc., we report the percentage of subjects who agreed with the machine at above-chance rates, and the percentage of images that showed above-chance human-machine agreement. For Experiment 1, the 95\% confidence interval for the \% of subjects with above-chance classification was [94.6\% 99.4\%], and it was [82.8\% 98.7\%] for the \% of images with above-chance classification. For Experiment 2, these intervals were [87.7\% 95.7\%] and [58.0\% 83.7\%], respectively. For Experiment 3, these intervals were [83.2\% 93.2\%] and [67.7\% 90.7\%], respectively. Across all three experiments, these outcomes were reliably different from chance at \textit{p}$<$.001 (two-sided binomial probability test).}
\label{fig:Figure 2}
\end{figure*}

\noindent On each trial, subjects (N=200) saw one fooling image, displayed above both its CNN-generated label and a label randomly drawn from the other 47 images. Subjects selected whichever of the two labels they thought the machine generated for that image (Fig.\ \ref{fig:Figure 2}a).\\
 
\noindent Remarkably, human observers strongly preferred the machine's chosen labels to the foil labels: Classification ``accuracy" (i.e., agreement with the machine's classification) was 74\%, well above chance accuracy of 50\% (95\% confidence interval: [72.9\%, 75.5\%]; two-sided binomial probability test: \textit{p}$<$.001). Perhaps more tellingly, 98\% of observers chose the machine's label at above-chance rates, suggesting surprisingly universal agreement with the machine's choices (Fig.\ \ref{fig:Figure 2}d, ``\% of subjects showing who agree with the machine''). Additionally, 94\% of the images showed above-chance human-machine agreement: Only 3 images out of all 48 had corresponding CNN-generated labels that humans tended to reject compared to a random label, and 45/48 had CNN-generated labels that the humans tended to prefer over a random label (Fig.\ \ref{fig:Figure 2}d, ``\% of images with human-machine agreement''). This initial result suggests that human observers can broadly distinguish the features CNNs use to classify fooling images as familiar objects.

\vspace*{-.3cm}

\subsection*{Experiment 2: 1st choice vs.\ 2nd choice}

\vspace*{0.13cm}

How deep does this ability run? Though subjects in Experiment 1 could distinguish the machine's chosen label from a random label, they may have achieved this reliable classification not by discerning any meaningful resemblance between the images and their CNN-generated labels, but instead by identifying very superficial commonalities between them (e.g., preferring \textit{bagel} to \textit{pinwheel} for an orange-yellow blob simply because bagels are also orange-yellow in color).\\

\noindent To ask whether humans can appreciate subtler distinctions made by the machine, Experiment 2 contrasted the CNN's preferred label not with random labels but instead with the machine's \textit{second}-ranked label for that image. For example, considering the round golden blob in Figure \ref{fig:Figure 2}, AlexNet's next choice after \textit{bagel} is \textit{pretzel}, which similarly implies a rounded golden object. So, we obtained these second-ranked choices for every fooling image, and asked observers in Experiment 2 to choose between the machine's first choice and the machine's second choice --- i.e., between \textit{bagel} and \textit{pretzel} for the golden blob image, and so on for all 48 images (Fig.\ \ref{fig:Figure 2}c).  \\

\vspace*{0.05cm}

\noindent Again, human observers agreed with the machine's classifications: 91\% of observers tended to choose the machine's 1st choice over its 2nd choice, and 71\% of the images showed human-machine agreement (Fig.\ \ref{fig:Figure 2}d). Evidently, humans can appreciate deeper features within adversarial images that distinguish the CNN's primary classification from closely competing alternatives. Moreover, this result also suggests that humans and machines exhibit overlap even in their rank-ordering of image labels, since Experiment 2 yielded less human-machine agreement than Experiment 1 (94\% of images vs. 71\% of images). This suggests that the CNN's second-choice was also moderately intuitive to human subjects --- more so than a random label, but less so than the machine's first-choice label, just as would be expected if machine and human classification were related in this way.

\vspace*{0.15cm}

\subsection*{Experiment 3a: Many-way classification}

\vspace*{0.25cm}

The above experiments show that humans can identify the machine's preferred label from among relevant alternatives. However, both of these studies involve the limited case of only two alternatives; by contrast, image-recognizing CNNs typically choose from hundreds or thousands of labels when classifying such images. Would humans exhibit reliable agreement with the machine even under more unconstrained circumstances? Although it would not be practically feasible to make humans choose from 1,000 individual labels, Experiment 3 stepped closer to these conditions by displaying the labels of all 48 images at once, and asking subjects to pick the best of all the labels for each image.  \\

\vspace*{0.06cm}

\noindent Even under these demanding conditions, 88\% of subjects selected the machine's label at above-chance rates, and 79\% of images showed above-chance human-machine agreement. Moreover, in an analysis inspired by the \textit{rank-5} measure in the machine-learning literature, we found that the machine's label was among the top five human choices for 63\% of the images (whereas random chance would put this figure at approximately 10\%); in other words, even when the single most popular human-chosen label was not the CNN's preferred label, the 2nd, 3rd, 4th, or 5th most popular human-chosen label (out of 48 possible choices) usually did match the CNN's preferred label. These results suggest that humans show general agreement with the machine even in the taxing and unnatural circumstance of choosing their classification from dozens of labels displayed simultaneously.\\

\vspace*{0.15cm}

\subsection*{Experiment 3b: ``What is this?''}

\vspace*{0.15cm}

The previous study more closely resembled the task faced by CNNs in classifying images, which is to classify an image from among many labels. However, all of the preceding experiments differ from a CNN's task in another way: Whereas CNNs select a label that best matches an image, our human subjects were asked to \textit{anticipate the machine's label}, rather than to label the images themselves. Would humans still agree with the CNN's classification if their task were simply to straightforwardly classify the image? \\

\noindent Experiment 3b investigated this question by changing the task instructions: Rather than being told to ``think like a machine'' and guess a machine's preferred label, subjects were simply shown images and asked ``What is this?''. On each trial, an image appeared on the display, and subjects were asked ``If you \textit{had} to pick a label for it, what would you pick?'', from 48 possible labels. Once again, human judgments and machine classifications converged: 90\% of subjects agreed with the machine at above-chance rates, and 81\% of the images showed above-chance human-machine agreement. These results suggest that the humans' ability to decipher adversarial images doesn't depend on the peculiarities of our \textit{machine-theory-of-mind} task, and that human performance reflects a more general agreement with machine (mis)classification.

\vspace*{0.15cm}

\subsection*{Experiment 4: Television static images}

\vspace*{0.25cm}

Though the images in the above experiments are peculiar, they do at least have discrete and distinguishable features; for example, the \textit{baseball} image has a collection of diagonally intersecting red lines that resemble the characteristic red stitching of a real baseball. (Indeed, the creators of this adversarial attack informally noted this resemblance in later work; Nguyen, Yosinski, \& Clune, 2016.) What about truly bizarre images that are considered ``totally unrecognizable to human eyes" (Nguyen et al., 2015, p.427)? \\

\vspace*{0.06cm}

\begin{figure*}[!ht]
\centering
\includegraphics[scale=0.17]{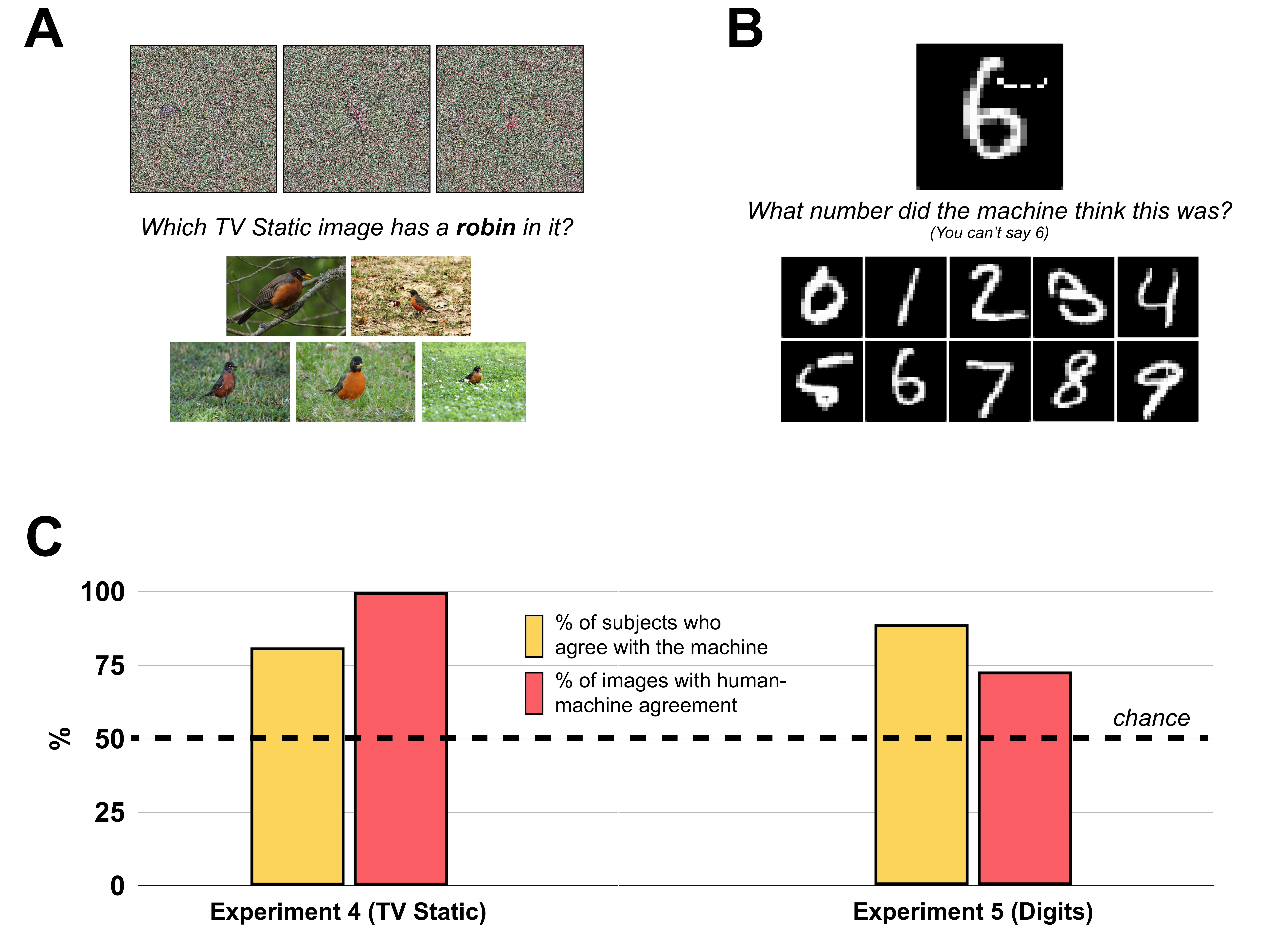}
\captionsetup{width=.85\linewidth,font=footnotesize}
\caption
{Classification with directly encoded fooling images and perturbed MNIST images. (A) In Experiment 4, 200 subjects saw eight directly encoded ``television static" images at once (though only three are displayed here); on each trial, a single label appeared, along with five natural photographs of the label randomly drawn from ImageNet (here, a centipede). The subjects' task was to pick whichever fooling image corresponded to the label. (B) In Experiment 5, 200 subjects saw 10 undistorted handwritten MNIST digits at once; on each trial, a single distorted MNIST digit appeared (100 images total). The subjects' task was to pick whichever of the undistorted digits corresponded to the distorted digit (aside from its original identity. (C) Most subjects agreed with the machine more often than would be predicted by chance responding, and most images showed human-machine agreement more often than would be predicted by chance responding (including every one of the television static images). For Experiment 4, the 95\% confidence interval for the \% of subjects with above-chance classification was [75.5\% 87.3\%], and [63.1\% 100\%] (one-sided 97.5\% confidence interval) for the \% of images with above-chance classification. For Experiment 5, these intervals were [84.2\% 93.8\%] and [64.3\% 81.7\%], respectively. Across both experiments, these outcomes were reliably different from chance at \textit{p}$<$.001 (two-sided binomial probability test).}

\vspace*{-0.1cm}
\label{fig:Figure 3}
\end{figure*}

\noindent In Experiment 4, subjects saw eight ``television static" images that CNNs recognize as objects --- e.g., \textit{centipede} or \textit{robin} (Fig.\ \ref{fig:Figure 3}a). These images appear to be colorful collections of pixels with little if any underlying structure. (However, upon \textit{very} close inspection, you may notice a small, often central, `object' within each image.) On each trial, a given label appeared on the screen, along with five examples of that category drawn from ImageNet (e.g., the word ``robin" beside five photographs of robins). Subjects were instructed to select the television-static image that best matched the label (Fig.\ \ref{fig:Figure 3}a).  \\

\vspace*{0.06cm}

\noindent Even with these bizarre images, 81\% of observers agreed with the machine at above-chance rates, and 100\% of the images showed above-chance human-machine agreement (i.e., they were chosen as matches more than 12.5\% of the time; Fig.\ \ref{fig:Figure 3}c). Moreover, for 75\% of the images, the label chosen most often by subjects was also the machine's most preferred choice (analogous to \textit{rank-1} performance). This is especially relevant for human-machine comparisons, since CNNs typically make their classification decisions after a softmax transformation has been applied to the input to the CNN's final layer; applying a similar transformation over our human responses could thus similarly produce `high confidence' ratings for adversarial images, if we were to treat our entire human cohort's judgments as `votes' over which a softmax decision is computed.  \\

\noindent These results suggest that human subjects are not only able to discern subtle features of adversarial images, but they also can infer machine classification of such images even when the relevant patterns are not discrete features at all but instead seemingly featureless collections of colored pixels.

\vspace*{0.2cm}

\subsection*{Experiment 5: Perturbed digits}

\vspace*{0.2cm}

The foregoing experiments explored ``fooling" images that human observers would not typically classify as familiar objects. However, a more insidious form of adversarial attack can occur when a few perturbed pixels fool CNNs into classifying one natural object as a \textit{different object}; for example, an image that would normally be classified as a ``4" might now be classified as a ``7" when just a small subset of the pixels is altered (Papernot et al., 2016; Fig.\ \ref{fig:Figure 1}c). This sort of attack is of special practical importance: One could imagine, for example, a malicious actor altering a speed limit sign in this way, which might fool an autonomous vehicle into recognizing a \textit{Speed Limit 45} sign as a \textit{Speed Limit 75} sign and then dangerously accelerating as a result.\\

\noindent The original research that generated such images concluded that ``humans cannot perceive the perturbation introduced to craft adversarial samples" (Papernot et al., 2016), because human observers persisted with their original classifications even after the distortion was introduced (see also Harding et al., 2018). By contrast, here we asked humans which digit they \textit{would} have picked if they weren't allowed to give their initial impression. We collected 100 adversarially distorted digits that had caused a CNN (LeNet; LeCun et al., 1998) to change its classification, and asked subjects which digit they thought the machine (mis)perceived the images as (Fig.\ \ref{fig:Figure 3}b).  \\

\noindent Even for perturbed adversarial images, human responses again aligned with the machine's: 89\% of subjects identified the machine's classifications at above-chance rates, and 73\% of images showed above-chance human-machine agreement (Fig.\ \ref{fig:Figure 3}c). Thus, even when adversarial images have strong prepotent identities, humans can anticipate the machine's misclassifications.

\vspace*{0.1cm}

\subsection*{Experiment 6: Natural images and localized perturbations}

\vspace*{0.2cm}

Whereas the previous result suggested that humans can decipher not only \textit{fooling} images but also \textit{perturbed} images, the particular adversarial attack explored in Experiment 5 may be limited in important ways: The proportion of perturbed pixels was often relatively high (as many as 14\% of the pixels in the image); the perturbations often obstructed salient parts of the image; and the target of the adversarial attack was only handwritten digits, which differ from natural images both in their general richness and also in the breadth of possible target classes (since they involve only the digits 0-9). By contrast, more recent adversarial attacks overcome many of these limitations; could humans decipher the images produced by more advanced approaches?\\

\vspace*{0.05cm}

\noindent Experiment 6 tested human observers on images produced by a state-of-the-art ``localized" adversarial attack (``LaVAN"; Karmon et al., 2018). This adversarial distortion perturbs far fewer pixels in the attacked image; it succeeds even when the perturbation is confined to an isolated corner of the image (rather than obstructing the image's focal object); and it can target a wider array of natural images. For example, LaVAN can cause a machine to misclassify a daisy as a jaguar or a subway train as a milk can (as in Fig.\ \ref{fig:Figure 4}a), and it can do so even while perturbing only 2\% of pixels near the border of the image. As the authors of that work note, however, the perturbed pixels sometimes look like miniature versions of the adversarial target classes. Do naive human subjects agree? \\

\vspace*{0.06cm}

\noindent We acquired 22 such images that caused a CNN (Inception V3; Szegedy et al., 2016) to misclassify, and placed them in the same forced-choice design as Experiment 1, but with the addition of natural images of the target and foil classes randomly taken from ImageNet (so that subjects viewing a subway-train-to-milk-can image, for example, saw the label ``milk can" along with five images of milk cans drawn from ImageNet, and similarly for the foil labels; Fig.\ \ref{fig:Figure 4}a). Even for this advanced adversarial attack, human responses aligned with the machine's: 87\% of subjects identified the machine's classifications at above-chance rates, and 100\% of the images showed above-chance human-machine agreement (Fig.\ \ref{fig:Figure 4}c). Thus, even more recent and sophisticated adversarial attacks are susceptible to human deciphering.

\vspace*{0.1cm}

\subsection*{Experiment 7: 3D objects}

\vspace*{0.2cm}

All of the preceding experiments explored adversarial attacks on 2D \textit{images}, including both natural photographs and digitally generated textures. Such images are, certainly, the predominant targets of adversarial attacks; however, these attacks are ultimately limited in (a) their complexity, since the images are only two-dimensional; (b) their practical applications, since they typically ``fool" image-recognizing computer systems only when those systems are fed such images directly; (c) their robustness, since most attacks on 2D images lose their fooling powers when the images are rotated, resized, blurred, or otherwise manipulated; and (d) their promise for understanding the richness of human object representation, since we typically see real-life objects in the world from multiple angles and with multiple cues, rather than a single image from a single viewpoint with only pictorial image cues.\\

\begin{figure*}[!ht]
\centering
\includegraphics[scale=0.165]{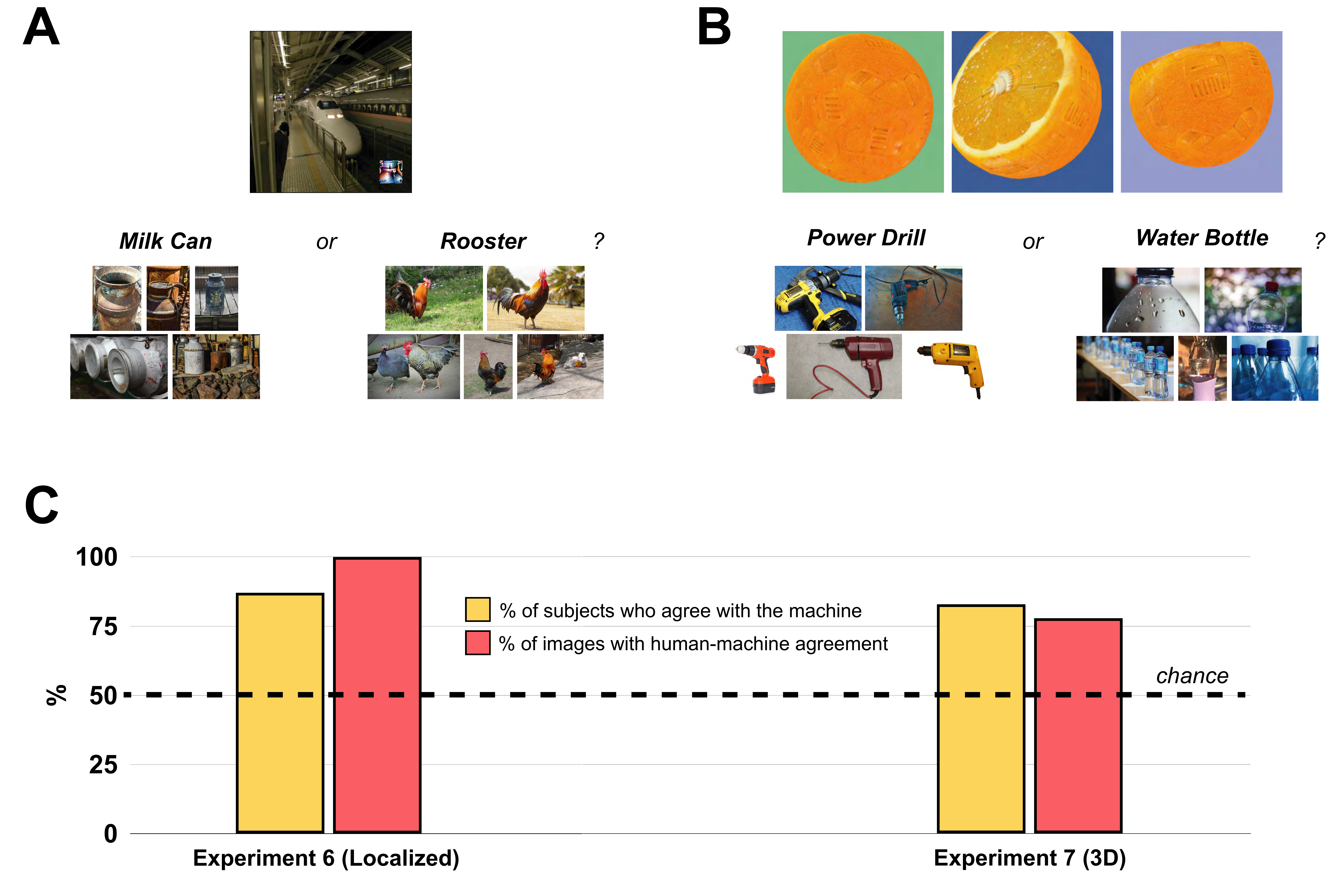}
\captionsetup{width=.85\linewidth,font=small}
\vspace*{-0.1cm}
\caption
{Classification with perturbed natural images and 3D objects. (A) In Experiment 6, 200 subjects saw natural photographs that had a small adversarial perturbation in the bottom right corner of the image (22 images total); the subjects' task was to choose between the machine's choice for that image and a random label drawn from the other images in the imageset (which were accompanied by images of each target class for reference). (B) In Experiment 7, 400 subjects saw three viewpoints of a rendered 3D object whose adversarial perturbation caused a machine to misclassify (106 images total); the subjects' task was to choose between the machine's choice for that object and a random label drawn from the other objects in the imageset (with images of each target class for reference). (C) Most subjects agreed with the machine more often than would be predicted by chance responding, and most images showed human-machine agreement more often than would be predicted by chance responding (including every one of the LaVAN images). For Experiment 6, the 95\% confidence interval for the \% of subjects with above-chance classification was [82.5\% 91.9\%], and [84.6\% 100\%] (one-sided 97.5\% confidence interval) for the \% of images with above-chance classification. For Experiment 7, these intervals were [78.7\% 86.5\%] and [70.5\% 86.1\%], respectively. Across both experiments, these outcomes were reliably different from chance at \textit{p}$<$.001 (two-sided binomial probability test).}

\vspace*{-0.2cm}
\label{fig:Figure 4}
\end{figure*}

\noindent Recently, a creative and sophisticated adversarial attack has allowed for the creation not only of adversarial 2D images but also adversarial 3D \textit{objects} that machines reliably misclassify (Athalye et al., 2017). When rendered using 3D graphics software, such ``robust" adversarial examples continue to fool CNN-based classifiers (here, Inception V3; Szegedy et al., 2016), not only from one particular vantage point but also from multiple different angles and distances; moreover, they can even be 3D printed as physical objects in the real world. A 3D model of an orange, for example, could be produced in physical form, placed on a table, and recognized by a machine as a power drill, cucumber, or even a \textit{missile}, simply because of certain vague textural elements on the orange's surface (Fig.\ \ref{fig:Figure 4}b). \\

\vspace*{-0.05cm}

\noindent Experiment 7 tested human observers on such robust 3D adversarial objects. We acquired 106 such examples, encompassing 10 familiar objects (e.g., baseball, turtle, orange) that are classified as something else (e.g., \textit{lizard}, \textit{puzzle}, \textit{drill}) when a certain texture is added to them. On each trial, human observers saw three different rendered viewpoints of each adversarial object, and were shown both the target label and a foil label drawn from another element in the imageset, with examples of the classes shown beneath (so that subjects viewing an orange-to-drill image, for example, saw the label ``power drill" along with five images of power drills drawn from ImageNet, and similarly for a foil label; \ref{fig:Figure 4}b). \\

\noindent Even for this state-of-the-art attack, human responses aligned with the machine's: 83\% of subjects identified the machine's classifications at above-chance rates, and 78\% of the images showed above-chance human-machine agreement (Fig.\ \ref{fig:Figure 4}c). Once again, humans were able to decipher the machine's classifications, here for one of the most advanced and alarming adversarial attacks in the literature. \\

\vspace*{-.22cm}

\section*{General Discussion}

The present results suggest that human intuition is a reliable source of information about how machines will classify images --- even for adversarial images that have been specifically designed to fool the machine. This implies at least some meaningful degree of similarity in the image features that humans and machines prioritize --- or \textit{can} prioritize --- when associating an image with a label. The very existence of adversarial images has cast into doubt whether recently developed machine-vision systems bear any real resemblance to humans in terms of how they classify images, and also whether such models can be attacked surreptitiously. The present results suggest that may indeed have such overlap, and perhaps even that humans could play a role in ``defending" against, or even further refining, such attacks. \\

\noindent How deep does this ability run? The human subjects here showed reliable agreement with the machine across an impressively broad array of images: collages of features, television-static images, handwritten digits, natural photographs, and 3D objects. There is also reason to think these abilities could generalize further. For example, recent work has shown that physically placing a small and colorful ``sticker" next to a banana can fool CNNs into classifying images of the banana as a toaster (Brown et al., 2017); however, the sticker itself \textit{looks} quite like a toaster, and we suspect that other attacks may be similarly decipherable.  \\

\noindent At the same time, there is a cottage industry around the production of adversarial images, and there may well be adversarial images that humans cannot decipher in the manner explored here. For example, some kinds of adversarial images are produced by making thousands of miniscule perturbations across every pixel in the image (as in a famous panda-to-gibbon example; Goodfellow, Shlens, \& Szegedy, 2014); we doubt humans could see a gibbon in this image, even under forced-choice conditions. However, there are at least two reasons why such images may actually be \textit{less} powerful as challenges to human-machine comparison than the images we explored here.  \\

\noindent \textit{First}, and more practically, those examples are the very cases that exhibit the least robust transfer across systems and transformations. For example, even a small rotation or rescaling of the perturbed image is usually sufficient to return it to its prior classification, which suggests that this is not the most practical attack for real-world settings. (For example, an autonomous vehicle that \textit{photographed} such images in the real world would almost certainly fail to be fooled; Lu et al., 2017.) Instead, the sort of adversarial attack that is more likely to succeed against an autonomous vehicle or baggage-screening operation is exactly the sort having some sort of visible noise pattern, and so that is the sort we explored here. \\

\noindent \textit{Second}, and more theoretically, the \textit{reason} such perturbations are not visible to humans may have little to do with the high-level processes underlying human object classification, but instead with low-level physiological limitations on human visual acuity, resolution, and sensitivity to contrast, which simply cannot match the power of \textit{in silica} image processing. In other words, it is plausible that humans cannot perceive certain perturbations simply because of the limitations of their eyes and early visual systems, rather than because of the concepts or templates they deploy in classifying objects. (Indeed, many instances of the panda-to-gibbon example \textit{must} should be undecipherable to humans because the perturbation in such images is often too small to change the value of any actual pixel as rendered on a monitor.) For this reason, the mere existence of such adversarial images perhaps tells us less about similarities or differences between humans and CNNs in high-level \textit{object classification} per se, but rather have to do with lower-level considerations such as the resolution of human vision or even of display equipment. (Similarly, some adversarial images  allow noisy pixels to take any value that the neural network can process, including those outside the dynamic range of images, as in other work by Karmon et al., 2018. Humans may have difficulty deciphering those patterns as well, but perhaps not because of the principles of human object recognition.) \\
	
\noindent To be sure, our results do not suggest that adversarial images are somehow unproblematic in the applied settings for which CNNs are hoped to be useful; adversarial images remain a dangerous and alarming development. But the present results do at least suggest that human intuition about such images can be a meaningful source of information about how a machine will classify them, and even that humans could have a role to play in the ``loop" that generates such images (Biggio \& Roli, 2018). For example, a small minority of the images in the present experiments (e.g., 3/48 in Experiment 1) had CNN-generated labels that were actively rejected by human subjects, who failed to pick the CNN's chosen label even compared to a random label drawn from the imageset. Such images better meet the ideal of an adversarial example, since the human subject actively rejects the CNN's label. However, we note that it was not clear \textit{in advance} of collecting the human data exactly which images the humans would be able to decipher and which they would not. An important question for future work will be whether adversarial attacks can ever be refined to produce \textit{only} those images that humans cannot decipher, or whether such attacks will always output a mix of human-classifiable and human-unclassifiable images; it may well be that human validation will always be required to produce such truly adversarial images (and that human testing on candidate adversarial images should be incorporated into the pipeline of testing and validating new CNN-based models of object classification). Indeed, one could state this possibility as a \textit{conjecture} (call it the ``knowable noise" conjecture): As long as (a) an adversarial attack produces noise that is visible, (b) the adversarial image is robust and transferable, and (c) the to-be-attacked system demonstrates human-level recognition accuracy on a wide array of images, that attack will tend to produce images that are judged by humans to resemble their target class.\\

\noindent A related question is whether human subjects could, with training, improve their ability to decipher adversarial images. For example, Experiment 2 (with ``television static" images) involved perhaps the most challenging and unfamiliar sorts of adversarial images, and exhibited a powerful practice effect, with a strong positive correlation between trial number and classification accuracy, evident even as a simple linear correlation, \textit{r}(6)=0.79, \textit{p}$<$.02. (This can also be shown by comparing classification accuracy on the final trial vs.\ the first trial, \textit{t}(166)=3.19, \textit{p}$<$.002). This suggests that greater familiarity with the space of adversarial images might allow humans to better anticipate the machine's classifications, and perhaps that future work could determine how best to prepare and train humans to detect and decipher such images. \\

\noindent What do these results say about the relationship between humans and machines? An important property of the adversarial examples studied here is that they were originally created without the human visual system in mind. Other work has produced images that cause humans to misclassify under choice- and time-limited circumstances (e.g., classifying an image of a distorted ``dog" as a ``cat" when the image is presented for 63ms; Elsayed et al., 2018). The conclusions of this work are consonant with our own, in that they show how humans and CNNs can be made to give similar classifications for adversarial images. However, one important difference is that the success of this earlier work required explicitly incorporating aspects of the human image-processing stream into the procedure for generating adversarial images. For example, the adversarial images produced by that procedure not only had to fool a CNN into misclassifying, but also had to first pass through models of the human retina and sophisticated forms of spatial blurring that incorporate real measurements from the primate visual system. By contrast, the images explored in the present studies were simply generated to fool a machine. In at least this sense, the present studies `stack the deck' against human-machine convergence, since the adversarial images we study here were generated without any consideration of human vision all; yet, we still find evidence for human deciphering of adversarial stimuli. \\

\noindent How, then, did our human subjects do this? As alluded to earlier, the answer may be in part that adversarial examples truly do share core visual features with the images they are mistaken for, especially considering the available labels. (Indeed, this simple fact may help explain why adversarial images generated for one CNN often transfer to other CNNs; Tram\`er et al., 2017). Why, then, does it seem so strange that such images should be classified as familiar objects? To be sure, it is unlikely that subjects in our experiments truly \textit{recognized} most of the adversarial images shown to them, in the sense of rapidly and spontaneously matching the image to a stored object representation; for example, it seems unlikely that humans could easily identify the adversarial images' target classes without at least some idea of the relevant label options. However, this possibility does not undermine the interest of the present results, for at least three reasons.\\

\noindent \textit{First}, even in the absence of spontaneous recognition, humans can engage in surprisingly sophisticated processing of even very sparse textures (Long, St\"ormer, \& Alvarez, 2017), and object identification in humans benefits in important and well-established ways from image labels and other contextual factors (Lupyan \& Thompson-Schill, 2012) --- especially for ambiguous or degraded images (Bar, 2004). For this reason, it is only natural that explicit labels and examples assist our human subjects, and that a given object identity isn't immediately forthcoming upon looking at the adversarial images. (Indeed, machine-vision systems themselves \textit{also} don't engage in ``free classification'' when they process adversarial images; they simply pick the best label in their provided vocabulary, just as our human subjects did. \textit{Second}, the real-world situations in which humans might one day encounter adversarial images may themselves involve known constraints on the relevant target classes: For example, if a human who sees some odd patterns on the number in a speed limit sign suspects it may be an adversarial image, the space of possible target classes may be fairly limited (as in Experiment 5). \textit{Third}, the distinctions made in cognitive science between rapid, effortless \textit{recognition} and slower, more deliberate reasoning simply do not exist for CNNs, whose architectures cannot easily be parsed in these ways. Though this very fact suggests an even deeper difference between humans and CNNs (for classic critiques of similar approaches, see Fodor \& Pylyshyn, 1988; Marcus, 1998), it also means that we cannot be sure that today's CNNs are doing \textit{recognition} either. In other words, even though our experiments may tap into human ``cognition" more than human ``perception" (Firestone \& Scholl, 2016), these distinctions may not even exist for CNNs --- and so both the CNNs' behavior and the humans' behavior might be readily interpreted as simply playing along with picking whichever label is most appropriate for an image.  \\

\noindent Indeed, although adversarial images are often analogized to optical illusions that flummox human vision (Kriegeskorte, 2015; Majaj \& Pelli, 2018; Yamins \& Dicarlo, 2016), we suggest another analogy: Whereas humans have separate concepts for appearing \textit{like} something vs.\ appearing \textit{to be} that thing --- as when a cloud looks like a dog without looking like it \textit{is} a dog, or a snakeskin shoe resembles a snake's features without appearing \textit{to be} a snake, or even a rubber duck shares appearances with the real thing without being \textit{confusable} for a duck --- CNNs are not permitted to make this distinction, instead being forced to play the game of picking whichever label in their repertoire best matches an image (as were the humans in our experiments). After all, the images in Figure \ref{fig:Figure 2}a \textit{do} look like pinwheels and bagels (at least, more than they look like baseballs or roofs); they just don't look like they \textit{are} pinwheels and bagels. Perhaps CNNs would agree, if they could. \\

\section*{Methods}

\subsection*{General Methods for Experiments 1-7}

\textit{Participants}. In all of the experiments reported here, separate groups of 200 subjects participated online through Amazon Mechanical Turk (for validation of this subject pool's reliability, see Crump et al., 2013). (In Experiment 7, 400 subjects participated, being randomly assigned to see one or another half of the images.) All groups of subjects (1800 total) provided informed consent and were compensated financially for their participation.\\

\noindent \textit{Procedure}. For the machine-theory-of-mind task, subjects were told about ``a machine that can look at a picture and tell us what it is", and also that the machine sometimes ``gives surprising answers". Subjects were told that the images that cause the surprising answers were collected here, and that their job was to guess what answer the machine gave (except in Experiment 3b, in which subjects were simply asked to classify the images). In Experiments 1, 2, 3a, 3b, 5, 6, and 7, adversarial images were displayed in different random orders for each subject (with each subject seeing each image exactly one time, except in Experiment 7, where each subject saw half of the images exactly one time each), and subjects clicked a button to indicate which label they thought the machine gave. (In Experiment 4, all 8 adversarial images were visible on every trial, and instead various candidate labels appeared in different random orders for each subject, with subjects picking the adversarial image that best matched the label shown on that trial.) The response options also appeared in random locations, with the machine's ``true" answer being equally likely to appear in any button location. After giving a response, the images and buttons disappeared for 500ms, after which the next trial appeared. In all experiments, subjects who quit early or otherwise failed to submit complete a dataset were excluded from further analysis, as were subjects whose median response time across all trials was less than 1000ms, which suggested that they simply clicked through the experiment without actively participating. Post-exclusion sample sizes for Experiments 1-7 were 185 (E1), 181 (E2), 161 (E3a), 174 (E3b), 167 (E4), 164 (E5), 195 (E6), and 368 (E7). However, no result reported here depended in any way on these exclusions; i.e., every pattern remained statistically reliable even without excluding any subjects.) \\

\noindent \textit{Adversarial stimuli}. Experiments 1-3 used a set of 48 indirectly encoded ``fooling" images obtained from Nguyen et al.\ (2015); Experiment 4 used 8 additional directly encoded images; Experiment 5 used 100 distorted images appearing in Papernot et al.\ (2016); Experiment 6 used 22 distorted images generated by Karmon et al.\ (2018) that resulted in a $>$75\%-confidence classification of the adversarial target class; and Experiment 7 used 106 images generated by Athalye et al.\ (2017) that resulted in a $>$95\%-confidence classification of the adversarial target class from at least 3 views (with the top 3 such views displayed to subjects on a given trial). All images appeared to subjects at their native resolution, without any additional compression or distortion (unless subjects actively zoomed their browser in or out). \\

\noindent \textit{Data availability}. All data, code, and materials that support the findings of this study are available at \url{https://osf.io/uknbh/?view_only=31b9c3bb5b924aecb721466fc6e6ebd8}.

\section*{Acknowledgments}

For helpful correspondence and for sharing images and details from their work, we thank Anish Athalye, Danny Karmon, Anh Nguyen, and Nicolas Papernot. For comments on earlier drafts, we thank Christopher Honey and Daniel Yamins. 

\section*{Author Contributions}

C.F. and Z.L. contributed to the design and execution of the experiments and data analyses, and C.F. and Z.L. wrote the paper in collaboration. The authors declare no competing interests.

\appendix

\end{document}